\documentclass[10pt,twocolumn,letterpaper]{article}

\usepackage{cvpr}
\usepackage{times}
\usepackage{epsfig}
\usepackage{graphicx}
\usepackage{amsmath}
\usepackage{amssymb}

\usepackage{tabularx}
\usepackage{booktabs}
\usepackage{multirow}

\usepackage{algorithmicx}

\usepackage{pifont} 
\newcommand{\cmark}{\ding{51}}%
\newcommand{\xmark}{\ding{55}}%

\def\cf{\emph{cf}\onedot} 

\usepackage[breaklinks=true,bookmarks=false]{hyperref} 

\cvprfinalcopy 

\def\cvprPaperID{1284} 

\ifcvprfinal\fi

\newcommand{\argmax}{\operatornamewithlimits{arg\,max}}

\begin{document}

\title{Action Sets: Weakly Supervised Action Segmentation without Ordering Constraints}

\author{Alexander Richard, Hilde Kuehne, Juergen Gall\\
University of Bonn, Germany\\
{\tt\small \{richard,kuehne,gall\}@iai.uni-bonn.de}
}

\maketitle

\begin{abstract}
Action detection and temporal segmentation of actions in videos are topics of
increasing interest. While fully supervised systems have gained much attention
lately, full annotation of each action within the video is
costly and impractical for large amounts of video data. Thus, weakly supervised
action detection and temporal segmentation methods are of great importance. While
most works in this area assume an ordered sequence of occurring actions to be
given, our approach only uses a set of actions. Such action sets provide much less
supervision since neither action ordering nor the number of action occurrences are
known. In exchange, they can be easily obtained, for instance, from meta-tags,
while ordered sequences still require human annotation.
We introduce a system that automatically learns to temporally
segment and label actions in a video, where the only supervision that is used are
action sets. An evaluation on three datasets shows that our method still achieves
good results although the amount of supervision is significantly smaller than for
other related methods.
\end{abstract}


\section{Introduction}
\label{sec:introduction}

\begin{figure*}
    \centering
    \includegraphics{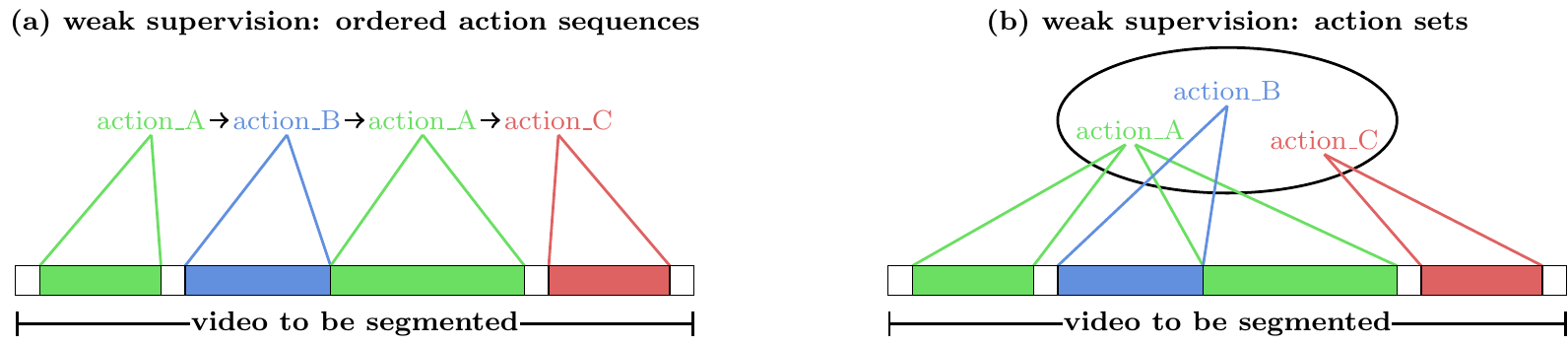}
    \caption{(a) Weak supervision with ordered action sequences
                  ~\cite{bojanowski2014weakly, kuehne2017weakly, huang2016connectionist}.
                   The number of actions and their ordering is known.
             (b) Weak supervision with action sets (our setup). Note that neither
                    action orderings nor the number of occurrences per action are provided.}
    \label{fig:task}
\end{figure*}

Due to the huge amount of publicly available video data, there is an increasing interest
in methods to analyze these data. In the field of human action recognition, considerable
advances have been made in recent years. A lot of research has been published on
action recognition, \ie action classification on pre-segmented video clips
\cite{wang2013action, simonyan2014two, karpathy2014large}. While current methods
already achieve high accuracies on large datasets such as UCF-101~\cite{soomro2012ucf101}
and HMDB-51~\cite{kuehne2011hmdb}, the assumption of having pre-segmented action
clips does not apply for most realistic tasks.
Therefore, there is a growing interest in efficient methods for finding actions
in temporally untrimmed videos. With the availability of large scale datasets
such as Thumos~\cite{thumos14}, Activity Net~\cite{caba2015activitynet}, or
Breakfast~\cite{kuehne2014breakfast}, many new approaches to temporally locate
and classify actions in untrimmed videos emerged~\cite{rohrbach2012database,
oneata2014lear, yeung2016endtoend, shou2016multistage, richard2016temporal, lea2016segmental}.
However, these approaches usually rely on fully supervised data, \ie the exact
temporal location of each action occurring in the training videos is known.
Creation of such training data requires manual annotation on video frame level
which is very expensive as well as impractical for large datasets.
Thus, there is a need for methods that can learn temporal action segmentation and
labeling with less supervision.
A commonly made assumption is that instead of full supervision, only an ordered
sequence of the actions occurring in the video is provided~\cite{bojanowski2014weakly,
kuehne2017weakly, huang2016connectionist, richard2017weakly}. Although this kind of weak supervision
is already much easier to obtain, \eg from movie scripts or subtitles, for a vast
amount of real world tasks, such information still can not be assumed to be
available. Instead, weak labels often arise in form of meta tags or unordered lists
from document indexing.

To address this problem, we propose a weakly supervised method that can learn temporal
action segmentation and labeling from unordered action labels, which we refer to as
action sets. In contrast to the above mentioned methods (\cf Figure~\ref{fig:task}a),
we assume that neither ordering nor number of occurrences of actions is provided during
training. Instead, only a set of actions occurring within the video is given
(\cf Figure~\ref{fig:task}b). 
This task is much more difficult than the case where ordered action transcripts are given.
Consider, for instance, a video with $ T $ frames
and a transcript of $ C $ ordered actions. Then, there are $ \frac{(C+T)!}{C!T!} $ possible
labelings for the video. If the actions are not ordered, there are already $ C^T $
possible labelings. For a very short video of $ 100 $ frames and $ C = 5 $, this means
that using unordered actions sets as supervision already allows for about $ 10^{60} $
times more possible labelings than when provided ordered action transcripts.

In order to deal with such an enormously large search space, we propose three model
components that aim at decomposing the search space on three different levels of
granularity. The coarsest level is addressed by a context model that
restricts the space of possible action sequences.
On a finer level, a length model restricts the durations of actions to a reasonable length.
On the lowest, most fine-grained level, a frame model provides class probabilities for each video frame.

Note that context models~\cite{kuehne2017weakly,richard2017weakly} and length models~\cite{richard2016temporal}
have been used before. However, in these works either ordered action transcripts for the
context model or framewise annotations for the length models are provided.
To the best of our knowledge, we are the first to use these models without being provided
any training data that allows to directly infer such models from the video annotation.

In an extensive evaluation, we investigate the impact of each component within the system.
Moreover, temporal segmentation and action labeling
quality is evaluated on unseen videos alone and on videos with action sets given
at inference time as additional supervision.


\section{Related Work}

Strong feature extractors developed in classical action recognition such as Fisher
vectors of improved dense trajectories \cite{wang2013action} or a variety of sophisticated
CNN methods~\cite{simonyan2014two, jain201515, karpathy2014large, feichtenhofer2016convolutional}
have also pushed the advances in untrimmed action segmentation.

When processing untrimmed videos, actions can either be localized in the temporal domain
only~\cite{bojanowski2014weakly,richard2016temporal,yeung2016endtoend,shou2016multistage,richard2017weakly,kuehne2017weakly,lea2017temporal,richard2018nnviterbi},
or in the spatio-temporal domain~\cite{vanGemert2015apt,jain2014action,mettes2016spoton,yan2017weakly}.
For the latter, videos are usually constrained to contain only few action instances.
While most approaches in this area are fully supervised, \cite{yan2017weakly} propose
a weakly supervised method for actor-action segmentation that is based on a multi-task
ranking model.

In this work, we focus on localizing actions in the temporal domain only.
In this setting, videos either contain multiple actions
of several classes occurring densely throughout the whole video~\cite{kuehne2014breakfast,
rohrbach2012database}, or sparsely~\cite{thumos14}, \ie most of the video is background and
all instances of a single class or a small set of classes need to be detected in the video.
Well studied methods from classical action recognition are frequently used as framewise
feature extractors~\cite{oneata2014lear, rohrbach2012database, yeung2016endtoend, richard2016temporal}.
Although CNN features are successful in some action detection methods~\cite{yeung2016endtoend, shou2016multistage},
they usually require to be retrained using full supervision. Improved dense trajectories, on
the contrary, are extracted in an unsupervised manner, making them the features of
choice for most weakly supervised approaches~\cite{bojanowski2014weakly, kuehne2017weakly,
huang2016connectionist, richard2017weakly}.

In the context of fully supervised action detection, most approaches use a sliding
window to efficiently segment a video~\cite{oneata2014lear, rohrbach2012database}
and rely on CNNs or recurrent networks~\cite{yeung2016endtoend,singh2016multistream,shou2016multistage}
that can not be used if only weak supervision is available.
The same holds for~\cite{richard2016temporal}, who
model context and length information, which is also done in our approach.
They show that length and context
information significantly improve action segmentation systems, using a Poisson
distribution to model action lengths and a language model to incorporate action
context information. Other fully supervised methods guided by grammars have been proposed
in~\cite{pirsiavash2014parsing,vo2014stochastic,kuehne2016end}.
Note that in contrast to our task, their length and context
model can be easily estimated from the frame-level training annotations. The challenge for
our problem formulation, however, is that no annotations that allow a direct estimation
of a context or length model are provided.

When working with weak supervision, existing methods use ordered action sequences
as annotation.
Early works suggest to get action sequences from movie scripts~\cite{laptev08learning,
duchenne2009automatic}. Alayrac \etal~\cite{alayrac2016unsupervised} propose to localize
specific actions in a video from narrated instructions.
In~\cite{malmaud15what}, it is proposed to use automatic speech recognition and align textual
descriptions, in their cases recipes, to the recognized spoken sequence.
Bojanowski \etal~\cite{bojanowski2014weakly} address the task of aligning actions to
frames. In their work, ordered action sequences are assumed to be provided during training
and testing and only an alignment between the frames and the action sequence is learned.
Kuehne \etal~\cite{kuehne2017weakly} extend their approach from~\cite{kuehne2016end}
to weak supervision by inferring a linear segmentation from ordered action sequences
and training a classical GMM+HMM speech recognition system on iteratively refined segmentations.
A further extension of this idea has been proposed by Richard \etal~\cite{richard2017weakly},
where the GMM is replaced by a recurrent neural network.
Recently, Huang \etal~\cite{huang2016connectionist} proposed to use connectionist temporal
classification (CTC) to learn temporal action segmentation from weakly supervised videos.
In order to avoid degenerate alignments between video frames and provided action transcripts,
they propose to use a visual similarity measure as an extension to the classical CTC approach.

In contrast to the approaches of~\cite{kuehne2017weakly, bojanowski2014weakly, huang2016connectionist, richard2017weakly},
our approach only uses action sets, \ie a much weaker supervision. Consequently, the way
our model is learned is also different from the above mentioned approaches.

Another recently published and related method by Wang \etal~\cite{wang2017untrimmed}
addresses the task of detecting an action in a video with sparse action occurrences.
More precisely, for a given action class, they generate action proposals and train
a neural network to distinguish instances of this action from background in the video.
Being designed to distinguish actions from background in a video, their method is
not suited for densely labeled videos containing many difference actions followed
by one another, as it is the case in this paper.


\section{Temporal Action Labeling}

\textbf{Task Definition.}
Let $ (x_1,\dots,x_T) $ be a video with $ T $ frames and $ x_t $
are the framewise feature vectors. The task is to assign an action label $ c $ from
a predefined set of possible labels $ \mathcal{C} $ to each frame of the video.
Following the notation of~\cite{richard2016temporal}, connected
frames of the same label can be interpreted as an action segment of class $ c $
and length $ l $. With this notation, the goal is to cut the video into an unknown
number of $ N $ action segments, \ie to define $ N $ segments with
lengths $ (l_1,\dots,l_N) $ and action labels $ (c_1,\dots,c_N) $.
To simplify notation, we abbreviate sequences of video frames, lengths, and classes
by $ \mathbf{x}_1^T $, $ \mathbf{l}_1^N $, and $ \mathbf{c}_1^N $, where the subscript
is the start index of the sequence and the superscript the ending index.

\textbf{Model Definition.}
In order to solve this task, we propose a probabilistic model and aim to find the most
likely segmentation and segment labeling of a given video,
\begin{align}
    (\mathbf{\hat l}_1^N, \mathbf{\hat c}_1^N) = \argmax_{N, \mathbf{l}_1^N, \mathbf{c}_1^N} \big\{ p(\mathbf{c}_1^N, \mathbf{l}_1^N | \mathbf{x}_1^T) \big\},
    \label{problem}
\end{align}
where $ l_n $ is the length of the $ n $-th segment and $ c_n $ is the corresponding
action label. We use a background class for all parts of the video in which no action
(or no action of interest) occurs. So, all video frames belong to one particular
action class and segment. Hence, $ \mathbf{l}_1^N $ and $ \mathbf{c}_1^N $ define
a segmentation and labeling of the complete video.

In order to build a probabilistic model, we first decompose Equation~\eqref{problem}
using Bayes rule,
\begin{align}
    (\mathbf{\hat l}_1^N, \mathbf{\hat c}_1^N) = \argmax_{N, \mathbf{l}_1^N, \mathbf{c}_1^N}
                               \big\{ p(\mathbf{c}_1^N) p(\mathbf{l}_1^N|\mathbf{c}_1^N) p(\mathbf{x}_1^T|\mathbf{c}_1^N,\mathbf{l}_1^N) \big\}.
    \label{bayes}
\end{align}
The first factor, $ p(\mathbf{c}_1^N) $ is the coarsest model, controlling the likelihood of action sequences.
The second factor, on a finer level, is a length model that controls the action durations,
and the third factor finally provides a likelihood of the video frames for a specific segmentation and labeling.
The same factorization has also been proposed in~\cite{richard2016temporal} for
fully supervised action detection.
We would like to emphasize that our model only shares the factorization with the
work of~\cite{richard2016temporal}. Due to weak supervision, the actual models we
use and the way they are trained are highly different.

\subsection{Weak Supervision}

While most works on weakly supervised temporal action segmentation
use ordered action sequences as supervision
\cite{huang2016connectionist, kuehne2017weakly, bojanowski2014weakly}, in our task,
only unordered sets of actions occurring in the video are provided, \cf Figure~\ref{fig:task}b.
Notably, neither the order of the actions nor the number of occurrences per action
is known. Assuming the training set consists of $ I $ videos, then the supervision
available for the $ i $-th video is a set $ \mathcal{A}_i \subseteq{\mathcal{C}} $ of
actions occurring in the video.

During inference, no action sets are provided for the video and the model has to infer
an action labeling from the video frames only.
As an additional task, we also discuss the case where action sets are given for
inference, see Section~\ref{sec:givenActions}.

In the following, the models for the three factors
$ p(\mathbf{c}_1^N) $, $ p(\mathbf{l}_1^N|\mathbf{c}_1^N) $, and
$ p(\mathbf{x}_1^T|\mathbf{c}_1^N,\mathbf{l}_1^N) $ from Equation~\eqref{bayes} are
introduced.

\subsection{Context Modeling with Context-free Grammars}
\label{sec:cfg}

Our first step to handle the huge search space is to restrict the
possible action orderings
using a context-free grammar $ \mathcal{G} $ in order to model
the context prior $ p(\mathbf{c}_1^N) $. Once the grammar is generated, define
\begin{align}
    p(\mathbf{c}_1^N) = \begin{cases}
                            \text{const}, & \text{if } \mathbf{c}_1^N \in \mathcal{G}, \\
                            0, & \text{otherwise.}
                        \end{cases}
\end{align}
Concerning the maximization in Equation~\eqref{bayes}, this means that each action
sequence generated by $ \mathcal{G} $ has the same probability and all other
sequences have zero probability, \ie they can not be inferred. We propose the following
strategies to obtain a grammar:

\textbf{Naive Grammar.} All action sequences that can be created using elements from
each action set from the training data are possible. Formally, this means
\begin{align}
    \mathcal{G}_\mathrm{naive} = \bigcup_{i=1}^I \mathcal{A}_i^*,
    \label{grammar_naive}
\end{align}
where $ i $ indicates the $ i $-th training sample and $ \mathcal{A}_i^* $ is the Kleene closure of $ \mathcal{A}_i $.

\textbf{Monte-Carlo Grammar.} We randomly generate a large amount of $ k $ action
sequences. Each sequence is generated by randomly choosing a training sample $ i \in \{1,\dots,I\} $.
Then, actions are uniformly drawn from the corresponding action set $ \mathcal{A}_i $ until
the accumulated estimated means $ \lambda_c $ of all drawn actions exceed the video length $ T_i $.
The mean lengths $ \lambda_c $ are estimated action class durations, see Section~\ref{sec:lengthmodel}.

\textbf{Text-Based Grammar.} Frequently, it is possible to obtain a grammar from external
text sources, \eg from web recipes or books. Given some natural language texts, we enhance the
monte-carlo grammar by mining frequent word combinations related to the action classes.
Consider two action classes $ v $ and $ w $, for instance \texttt{butter\_pan} and \texttt{crack\_egg}.
If either of the words \texttt{butter} or \texttt{pan} is preceding \texttt{crack} or \texttt{egg} in the
textual source, we increase the count $ N(v,w) $ by one. This way, word conditional probabilities
\begin{align}
    p(w|v) = \frac{N(v,w)}{\sum_{\tilde w} N(v,\tilde w)}
\end{align}
are obtained that have a high value if $ v $ precedes $ w $ frequently and a low value otherwise.
The actual construction of the grammar follows the same protocol as the monte-carlo grammar with the
only difference that the actions are not drawn uniformly from the action set but according
to the distribution $ p(w|v) $, where $ v $ is the previously drawn action class.

\subsection{Length Model from Action Sets}
\label{sec:lengthmodel}

While a grammar already introduces some ordering constraints,
the search space is still tremendously large, considering that actions
can be of arbitrary and even practically unreasonable durations. Therefore,
as a second step, we estimate a length model out of the scarce information
we get from the training data.
In order to model the length factor $ p(\mathbf{l}_1^N|\mathbf{c}_1^N) $, we assume conditional
independence of each segment length and further drop the dependence of all
class labels but the one of the current segment, \ie
\begin{align}
    p(\mathbf{l}_1^N|\mathbf{c}_1^N) = \prod_{n=1}^N p(l_n|c_n).
\end{align}
Each class-conditional $ p(l|c) $ is modeled with a Poisson distribution for
class $ c $.

For the estimation of the class-wise Poisson distributions, only the action
sets $ \mathcal{A}_i $ provided in the training data can be used. Ideally, the
free parameter of a Poisson distribution, $ \lambda_c $, should be set to the mean
length of action class $ c $. Since this can not be estimated from the action sets,
we propose two strategies to approximate the mean duration of each action class.

\textbf{Naive Approach.} In the naive approach, the frames of each training video
are assumed to be uniformly distributed among the actions in the respective action set.
The average length per class can then be computed as
\begin{align}
    \lambda_c = \frac{1}{|\mathcal{I}_c|} \sum_{i \in \mathcal{I}_c} \frac{T_i}{|\mathcal{A}_i|},
\end{align}
where $ \mathcal{I}_c = \{i: c \in \mathcal{A}_i\} $ and $ T_i $ is the length of the $ i $-th video.

\textbf{Loss-based.} The drawback of the naive approach is that actions that are usually
short are assumed to be longer if the video is long.
Instead, we propose to estimate the mean of all classes together.
This can be accomplished by minimizing a quadratic loss function,
\begin{align}
    \sum_{i=1}^I \sum_{c \in \mathcal{A}_i} (\lambda_c - T_i)^2 \quad \text{ subject to } \lambda_c > l_{\mathrm{min}},
    \label{lossBasedMean}
\end{align}
where $ l_{\mathrm{min}} $ is a minimal action length.
For minimization, we use constrained optimization by linear approximation (COBYLA)~\cite{powell1994direct}.

Note that the true mean length of action $ c $ is likely to be smaller than $ \lambda_c $ since
actions may occur multiple times in a video. However, this can not be included into the length
model since the action sets do not provide such information.

\subsection{Multi-task Learning of Action Frames}
\label{sec:network}

Given the grammar and the length model that already strongly restrict
the search space, the last missing factor is the actual framewise model
providing a likelihood for each class to be present in a given frame.

In order to model this last factor from Equation~\eqref{bayes},
we train a network with $ |\mathcal{C}| $ many binary softmax output layers.
Each layer predicts if for a given frame $ x_t $ label $ c $ is present,
\ie if $ c \in \mathcal{A}_i $ or not.
Since an action $ c $ usually occurs in different context, all frames belonging to
class $ c $ are always labeled with its true class $ c $ and some varying other classes.
Thus, a classifier can learn a strong response on the presence of the correct class
and weaker responses on the presence of other falsely assigned classes.
As loss of our network, we therefore use the accumulated cross-entropy loss of each
binary classification task.

In order to use the output probabilities of the multi-task network during inference,
they need to be transformed to model the last factor from Equation~\eqref{bayes},
$ p(\mathbf{x}_1^T|\mathbf{c}_1^N,\mathbf{l}_1^N) $. We therefore define the class-posterior
probabilities
\begin{align}
    p(c|x_t) := \frac{p(c \textit{ present} | x_t)}
                    {\sum_{\tilde c} p(\tilde c \textit{ present} | x_t) }
\end{align}
and transform them into class-conditional probabilities
\begin{align}
    p(x_t|c) \propto \frac{p(c|x_t)}{p(c)}.
\end{align}
Since the network is a framewise model, $ p(c) $ is also a framewise prior. More
specifically, if $ \mathrm{count}(c) $ is the total number of frames labeled with
\textit{$c$ present}, then $ p(c) $ is the relative frequency
$ \mathrm{count}(c) / \sum_{\tilde c} \mathrm{count}(\tilde c) $.

Assuming conditional independence of the video frames, the probability of an action
segment ranging from frame $ t_s $ to $ t_e $ can then be modeled as
\begin{align}
    p(\mathbf{x}_{t_s}^{t_e} | c) = \prod_{t=t_s}^{t_e} p(x_t|c).
    \label{segmentProb}
\end{align}
Framewise conditional independence is a commonly made assumption in multiple
action detection and temporal segmentation methods~\cite{richard2016temporal,kuehne2017weakly,kuehne2016end}.
Note that $ t_s $ and $ t_e $ are implicitly given by the segment lengths $ \mathbf{l}_1^N $.
For the $ n $-th segment in the video, $ t_s^{(n)} = 1 + \sum_{i<n} l_i $ and
$ t_e^{(n)} = \sum_{i\leq n} l_i $.

The third factor of Equation~\eqref{bayes} is now modeled using the previously
defined segment probabilities,
\begin{align}
    p(\mathbf{x}_1^T|\mathbf{c}_1^N,\mathbf{l}_1^N) := \prod_{n=1}^N p(\mathbf{x}_{t_s^{(n)}}^{t_e^{(n)}} | c_n).
\end{align}

\subsection{Inference}
\label{sec:inf}

With the explicit models for each factor, the optimization problem from Equation~\eqref{bayes}
reduces to
\begin{align}
    (\mathbf{\hat l}_1^N, \mathbf{\hat c}_1^N) = \argmax_{N, \mathbf{l}_1^N, \mathbf{c}_1^N \in \mathcal{G}}
                               \big\{ \prod_{n=1}^N p(l_n|c_n) \cdot p(\mathbf{x}_{t_s^{(n)}}^{t_e^{(n)}} | c_n) \big\}.
    \label{finalModel}
\end{align}
Note that the $ \argmax $ is only taken over action sequences that can be generated
by the grammar. Since the same probability has been assigned to all those sequences,
the factor $ p(\mathbf{c}_1^N) $ from Equation~\eqref{bayes} is a constant.
Moreover, the length model $ p(l_n|c_n) $ strongly penalizes unlikely action
durations and allows for an efficient pruning of unlikely segmentations.
Both together lead to a significant reduction of the search space.

The solution to Equation~\eqref{finalModel} can now be efficiently computed using a
Viterbi algorithm over context-free grammars, as widely used in automatic speech
recognition, see for example~\cite{jurafsky1995using}. The algorithm is linear
in the number of frames and therefore allows for efficient processing of videos
with arbitrary length. The authors of~\cite{richard2016temporal} have shown that
adding a length model increases the complexity from $ \mathcal{O}(T) $ to  $\mathcal{O}(TL) $,
where $ L $ is the maximal action length that can occur. In theory, there is no
limitation on the duration of actions, so inference would be quadratic in the number
of frames. In practice, however, it is usually possible to limit the maximal allowed
action length $ L $ to some reasonable constant, maintaining linear runtime.\footnote{Source code and details on the dynamic programming equations can be found on \url{https://alexanderrichard.github.io}}


\section{Experiments}
\label{sec:experiments}

In this section, we analyze the components of our approach, starting
with the grammar (Section~\ref{sec:grammar_eval}) and the length model (Section~\ref{sec:length_eval}),
before we compare our system to existing methods that use more supervision
(Section~\ref{sec:compSota}).

\subsection{Setup}

\textbf{Datasets.}
We evaluate our approach on three datasets for weakly supervised temporal
action segmentation and labeling, namely the Breakfast dataset~\cite{kuehne2014breakfast},
MPII Cooking 2~\cite{rohrbach2016cooking2}, and Hollywood Extended~\cite{bojanowski2014weakly}.

The \textbf{Breakfast} dataset is a large scale dataset comprising $ 1,712 $
videos, corresponding to roughly $ 67 $ hours of video and $ 3.6 $ million
frames. Each video is labeled by one of the $ 10 $ coarse breakfast related activities like
\textit{coffee} or \textit{fried eggs}. Additionally, a finer
action segmentation into $ 48 $ classes is provided which is usually used for
action detection and segmentation. Overall, there are nearly $ 12,000 $ instances
of these fine grained action classes with durations between a few seconds and
several minutes, making the dataset very challenging. The actions are densely
annotated and only $ 7\% $ of the frames are background frames. We use four splits
as suggested in~\cite{kuehne2014breakfast} and provide frame accuracy as evaluation
metric.

\textbf{MPII Cooking 2} consists of $ 273 $ videos with $ 2.8 $ million frames.
We use the $ 67 $ action classes without object annotations. Overall, around
$ 14,000 $ action segments are annotated in the dataset. The dataset provides a
fixed split into a train and test set, separating $ 220 $ videos for training.
With $ 29\% $, the background portion in this dataset is at a medium level.
For evaluation, we use the midpoint hit criterion as proposed in~\cite{rohrbach2012database}.

\textbf{Hollywood Extended} is a smaller dataset comprising $ 937 $ videos with roughly $ 800,000 $ frames.
There are about $ 2,400 $ non-background action instances from $ 16 $ different
classes. With $ 61\% $ of the frames, the background portion within this dataset
is comparably large. We follow the suggestion of~\cite{bojanowski2014weakly} and
use a $ 10 $-fold cross-validation. The originally proposed evaluation metric is
a variant of the Jaccard index, intersection over detection, which is only reasonable for a transcript-to-video alignment task where the transcripts and thus the action orderings are known for the test sequences
as in~\cite{bojanowski2014weakly} and \cite{huang2016connectionist}. 
For temporal action segmentation, only a video is given during inference and the number of predicted segments can differ from the number of annotated segments. In this case, the metric can not be used.          
Thus, we stick to the Jaccard index (intersection over union), which
is widely used in the domain of action detection~\cite{richard2016temporal, thumos14}
and has also been used on this dataset by~\cite{kuehne2017weakly}.

\textbf{Feature extraction.}
For a fair comparison, we use the same features as~\cite{kuehne2017weakly} and \cite{huang2016connectionist}.
Fisher vectors of improved dense trajectories~\cite{wang2013action} are extracted
for each frame and the result is projected to a $ 64 $-dimensional subspace using
PCA as proposed by Kuehne \etal~\cite{kuehne2016end}. Then, the features are
normalized to have zero mean and unit variance along each dimension.
If not mentioned otherwise, we use the monte-carlo grammar
and the loss-based length model. The in-depth evaluation of our approach is conducted
on Breakfast, final results on other datasets are reported in Section~\ref{sec:compSota}.

\textbf{Model.}
For the neural network in the framewise model
we use a simple feed forward network with a single hidden layer of $ 256 $
rectified linear units. Experiments with deeper models could not generalize
to the test data (VGG-16: accuracy of $ 0.031 $ on Breakfast).
We also evaluated the neural network based multiple instance learning approach
of~\cite{wu2015mil}, which also was not able to make reliable predictions (accuracy $ 0.089 $ 
on Breakfast). We therefore found the multi-task network as proposed in Section~\ref{sec:network}
to be a simple yet effective model.

\textbf{Efficient inference.}
During inference, we allow to hypothesize new segments only every $ 30 $ frames.
This allows for inference roughly in realtime without affecting the performance
of the system compared to a more fine-grained segment hypothesis generation.

\subsection{Effect of the Grammar}
\label{sec:grammar_eval}

%

\begin{table}[t]
    \centering
    \footnotesize
    \begin{tabularx}{0.48\textwidth}{lXrXr}
        \toprule
                             & & \multicolumn{3}{c}{frame accuracy} \\
                                           \cmidrule(lr){3-5}
        Grammar              & & train     & & test \\
        \midrule
        none                 & & $ 0.147 $ & & $ 0.099 $ \\
        naive                & & $ 0.194 $ & & $ 0.134 $ \\
        monte-carlo          & & $ 0.282 $ & & $ 0.233 $ \\
        \midrule
        manually created     & & $ 0.333 $ & & $ 0.269 $ \\
        ground truth         & & $ 0.367 $ & & $ 0.294 $ \\
        \bottomrule
    \end{tabularx}
    \caption{Evaluation of our method on Breakfast using different context-free grammars.
             As length model, the loss-based approach is used.}
    \label{tab:grammars}
    \vspace{-0.5cm}
\end{table}

The main contribution of the grammar is to limit the search space and remove unrealistic
action sequences.
We compare different kinds of grammars and report the frame accuracy on both,
test and train set. Recall that due to weak supervision, our method does not necessarily
provide good results on the training videos, making it interesting to investigate both sets.
As shown in Table~\ref{tab:grammars}, the use of a sophisticated grammar is
crucial for good performance. The naive grammar is only slightly better than the
system without any grammar. The monte-carlo grammar boosts the frame accuracy by $ 10\% $ on
the test set. Note that we found the number of $ k $ monte-carlo samples for the grammar
not to be critical and chose $ 1,000 $ randomly generated sequences for all experiments.
Using a ground truth grammar, \ie a grammar learned from ordered action transcripts (which are
not provided in our setting) gives an upper bound on the performance that can be reached by
improving the grammar only. Notably, the monte-carlo grammar is only $ 6\% $ below this upper bound.

For a further comparison,
we gave all action sets from the training data to an annotator who was asked to manually create
an ordered action sequence for each set. This manually created grammar serves as a comparison
of the purely data driven monte-carlo grammar to human knowledge. Although the manual grammar is
better, the frame accuracy only differs by $ 3.6\% $. Since the annotator on average only needed
one minute per action set, a manual grammar is also a cheap opportunity to add human knowledge
without the need to actually annotate videos.

\begin{table}[t]
    \centering
    \footnotesize
    \begin{tabularx}{0.48\textwidth}{Xrrr}
        \toprule
                                               & Breakfast           & Cooking 2             & Holl. Ext.  \\
                                               & \textit{frame acc.} & \textit{midpoint hit} & \textit{jacc. idx} \\
        \midrule
        monte-carlo                            & $ 0.233 $           & $ 0.098 $             & $ 0.093 $   \\
        text-based                             & $ 0.232 $           & $ 0.106 $             & $ 0.092 $   \\
        \bottomrule
    \end{tabularx}
    \caption{Evaluation of the text-based grammar. For Cooking 2, where the text sources are closely related
             to the content of the videos, an improvement can be observed.}
    \label{tab:text_based_grammar}
    \vspace{-0.3cm}
\end{table}

As proposed in Section~\ref{sec:cfg}, textual sources can be used to enhance the monte-carlo grammar by
restricting the transition between action classes to only the likely ones. We evaluate such a text-based
grammar for all three datasets. For Breakfast, we used a webcrawler to download more than $ 1,200 $ breakfast
related recipes, for Hollywood Extended, $ 10 $ movie scripts of IMDB top-ranked movies have been
downloaded, and for Cooking 2, we used the scripts provided by the authors of the dataset. These scripts
were obtained by asking annotators to write sequential instructions on how to execute the respective
kitchen task. Consequently, the text sources used for Breakfast and Hollywood Extended are only loosely
connected to the datasets, whereas the textual source for Cooking 2 covers exactly the same domain as
the videos. Not surprisingly, we find that only for this case, the text-based grammar leads to an improvement
over the monte-carlo grammar, \cf Table~\ref{tab:text_based_grammar}. For the other datasets, neither an
improvement nor a degradation is observed.

\subsection{Effect of the Length Model}
\label{sec:length_eval}

\begin{table}[t]
    \centering
    \footnotesize
    \begin{tabularx}{0.48\textwidth}{lXrXr}
        \toprule
                             & & \multicolumn{3}{c}{frame accuracy} \\
                                           \cmidrule(lr){3-5}
        Length model         & & train     & & test \\
        \midrule
        naive                & & $ 0.254 $ & & $ 0.201 $ \\
        loss-based           & & $ 0.282 $ & & $ 0.233 $ \\
        ground truth         & & $ 0.341 $ & & $ 0.257 $ \\
        \bottomrule
    \end{tabularx}
    \caption{Evaluation of our method on Breakfast using different length models.
             As grammar, the monte-carlo approach is used.}
    \label{tab:length_models}
    \vspace{-0.3cm}
\end{table}

Besides the choice of the context-free grammar, the length model is a crucial component
of our system. The estimated mean action lengths influence the performance in two ways:
first, they define the Poisson distribution that contributes to the actual length of hypothesized
action segments. Secondly, they have a huge impact on the number of action instances that are
generated for each action sequence in the monte-carlo grammar.

\textbf{Mean Length Approximation.}
We compare the two proposed mean approximation strategies, naive and loss-based
mean approximation, with a ground truth model, \ie the true action means estimated on a frame-level
ground truth annotation of the training data. The results are shown in Table~\ref{tab:length_models}.
The naive mean approximation suffers from some conceptual drawbacks. Due to
the uniform distribution of video frames among all actions occurring in the video, short actions
may be assigned a reasonable length as long as the video is also short. If the video is long, however,
short actions get the same share of frames as long actions, resulting in an
over-estimation of the mean for short actions and an under-estimation of the mean for long actions.
The loss-based mean approximation, on the contrary, can provide more realistic estimates by minimizing
Equation~\eqref{lossBasedMean}. Note that the solution of the problem in principle would allow for
negative action means. Hence, setting the minimal action length $ l_\mathrm{min} > 0 $ is crucial. In
practice, we want to ensure a reasonable minimum length and set $ l_\mathrm{min} = 50 $ frames, corresponding
to roughly two seconds of video. The loss-based mean approximation performs significantly better than
the naive approximation, increasing the frame accuracy by $ 3\% $.

Comparing these numbers to the ground truth length model reveals that particularly on the train set,
on which the ground truth lengths have been estimated, there is still room for improvement. Considering
the small amount of supervision that we can utilize to estimate mean lengths, \ie actions sets only,
and the small gap between the loss-based approach and the ground truth model on the test set, on the other
hand, we find that our loss-based method already yields a good approximation.

\textbf{Evaluating Different Length Models.}
\begin{figure}
    \centering
    \includegraphics{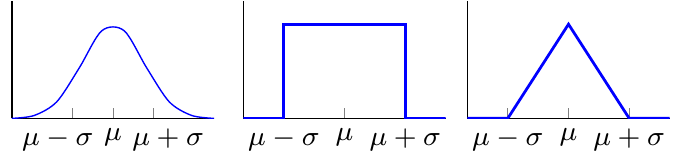}
    \footnotesize
    \begin{tabularx}{0.48\textwidth}{Xrrrr}
        \toprule
                 & Gaussian  & Box       & Triangle  & Poisson  \\
        \midrule
        accuracy & $ 0.148 $ & $ 0.220 $ & $ 0.227 $ & $ 0.233 $ \\
        \bottomrule
    \end{tabularx}
    \caption{Evaluation of different length models on Breakfast.}
    \label{fig:lengthmodels}
    \vspace{-0.5cm}
\end{figure}
So far we modeled the length with a Poisson distribution. There is a variety of other possible length
models. In Figure~\ref{fig:lengthmodels}, three additional models are evaluated,
a Gaussian, a box-, and a triangle model. Box and triangle model are zero outside $ [\mu - \sigma, \mu + \sigma] $.
The standard deviation $ \sigma $ of each action class is heuristically estimated by mapping actions according to their mean length
onto the possible segmentations generated by the monte-carlo grammar. The Gaussian model decays too fast around the mean lengths
and leads to low accuracies. Although the other models perform well, the Poisson distribution still yields the
best results.

\subsection{Impact of Model Components}

\begin{table}[t]
    \centering
    \footnotesize
    \begin{tabularx}{0.48\textwidth}{ccXrXr}
        \toprule
                &              & & \multicolumn{3}{c}{frame accuracy} \\
                                   \cmidrule(lr){4-6}
        grammar & length model & & train     & & test \\
        \midrule
        \xmark  & \xmark       & & $ 0.118 $ & & $ 0.080 $ \\
        \xmark  & \cmark       & & $ 0.147 $ & & $ 0.099 $ \\
        \cmark  & \xmark       & & $ 0.208 $ & & $ 0.154 $ \\
        \cmark  & \cmark       & & $ 0.282 $ & & $ 0.233 $ \\
        \midrule
        \multicolumn{2}{l}{fully supervised} & & $ 0.774 $ & & $ 0.556 $ \\
        \bottomrule
    \end{tabularx}
    \caption{The first four rows are a comparison of the impact of the grammar and the length model on the Breakfast dataset;
             the last is our system trained on fully supervised, \ie framewise annotated, data. It is an upper bound for the weakly supervised setup.}
    \label{tab:components}
    \vspace{-0.3cm}
\end{table}

\begin{figure}[t]
    \centering
    \includegraphics{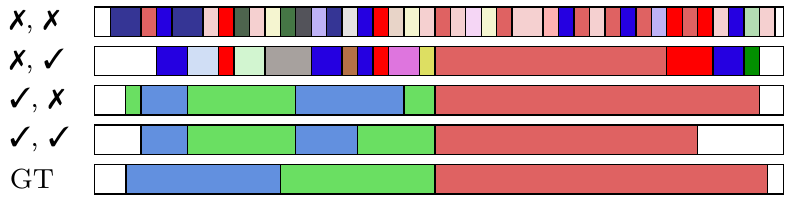}
    \caption{Example segmentation on a test video from Breakfast. Row one to four correspond to row
             one to four from Table~\ref{tab:components}. The last row is the ground truth segmentation.}
    \label{fig:segmentation}
    \vspace{-0.5cm}
\end{figure}

All three components, the grammar, the length model, and the framewise model,
contribute their share to restricting the search space to reasonable segmentations.
In this section, we evaluate the impact of the grammar and length model on their own
and in combination with each other.
We use the best-working
grammar and length approximation, \ie the monte-carlo grammar with loss-based mean
approximation, and analyze the effect of omitting the grammar and/or the length model
from Equation~\eqref{finalModel} during inference. The results are reported in Table~\ref{tab:components}.
Not surprisingly, the performance without a grammar is poor, as the model easily
hypothesizes unreasonable action sequences. Adding a grammar alone already boosts
the performance, restricting the search space to more reasonable sequences. In order
to also get action segments of reasonable length, however, the combination of grammar
and length model is crucial. This effect can also be observed in a qualitative segmentation
result, see Figure~\ref{fig:segmentation}. Note the strong over-segmentation if neither
grammar nor length model is used. Introducing the length model partially improves the
result but still the grammar is crucial for a reasonable segmentation in terms of
correct segment labeling and segment lengths.
The fully supervised model (last row of Table~\ref{tab:components}) is trained by assigning
the ground truth action label to each video frame. Apart from the labeling, the multi-task
network architecture remains unchanged. The full supervision defines an upper bound for our
weakly supervised method.

\subsection{Comparison to State of the Art}
\label{sec:compSota}

\begin{table}[t]
    \centering
    \footnotesize
    \begin{tabularx}{0.48\textwidth}{Xrrr}
        \toprule
                                               & Breakfast           & Cooking 2             & Holl. Ext. \\
                                               & \textit{frame acc.} & \textit{midpoint hit} & \textit{jacc. idx} \\
        \midrule
        \multicolumn{4}{l}{\textit{Weak supervision: unordered action sets}} \\
        monte-carlo                            & $ 0.233 $  & $ 0.098 $             & $ 0.093 $   \\
        text-based                             & $ 0.232 $           & $ 0.106 $    & $ 0.092 $            \\
        \midrule
        \multicolumn{4}{l}{\textit{Stronger supervision: ordered action transcripts}} \\
        HMM~\cite{kuehne2017weakly}            & $ 0.259 $           & $ 0.200 $             & $ 0.086 $   \\
        CTC~\cite{huang2016connectionist}      & $ 0.218 $           & $ - $                 & $ - $       \\
        ECTC~\cite{huang2016connectionist}     & $ 0.277 $           & $ - $                 & $ - $       \\
        HMM+RNN~\cite{richard2017weakly}       & $ 0.333 $           & $ - $                 & $ 0.119 $   \\
        \bottomrule
    \end{tabularx}
    \caption{Performance of our method compared to state of the art methods for weakly supervised
             temporal segmentation. Note that our method uses action sets as weak supervision,
             whereas~\cite{kuehne2017weakly, huang2016connectionist, richard2017weakly} have a stronger
             supervision with ordered action sequences.}
    \label{tab:sota}
    \vspace{-0.3cm}
\end{table}

\begin{figure}
    \centering
    \includegraphics{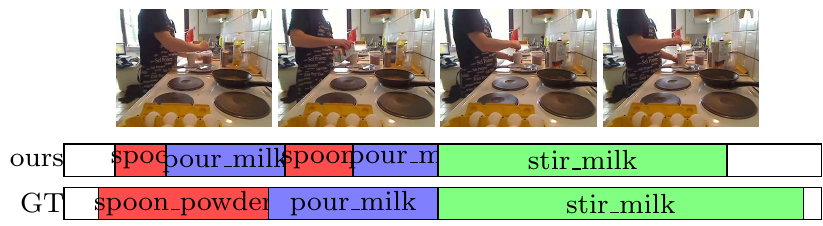}
    \caption{Example segmentation. All relevant ground truth actions are present. Note that \textit{spoon\_powder} always occurs jointly with \textit{pour\_milk},
             so it is hard for our model to distinguish them.}
    \label{fig:example}
\end{figure}

The task of weakly supervised learning of a model for temporal action segmentation given only action sets
has not been addressed before. Still, there are some works on temporal action segmentation
given ordered action sequences. In this section, we compare our approach to these
methods on the three datasets.
Kuehne \etal~\cite{kuehne2017weakly} approach the problem with hidden Markov models
and Gaussian mixture models and Richard \etal~\cite{richard2017weakly} extend their approach
using recurrent neural networks.
Huang \etal~\cite{huang2016connectionist}, in contrast, rely on connectionist
temporal classification (CTC) with LSTMs and extend it by downweighting degenerated
alignments and incorporating visual similarity of frames into the decoding algorithm.
They call their approach extended CTC (ECTC).
All of these approaches use ordered action sequences, and thus a much stronger
supervision than our method.
Keeping the tremendously large search space for our problem compared to~\cite{kuehne2017weakly,huang2016connectionist,richard2017weakly}
in mind (\cf Section~\ref{sec:introduction}), our model achieves remarkable results on Breakfast and Hollywood Extended, \cf Table~\ref{tab:sota}.
Note that training the HMM approach of~\cite{kuehne2017weakly} with monte-carlo sampled action transcripts
(\ie with the same amount of supervision as in this paper)
only yields an accuracy of $ 0.145 $ on Breakfast, which is far less than our approach.
An example segmentation of our approach is shown in Figure~\ref{fig:example}.
Falsely recognized actions are frequently those that only occur jointly, such as \texttt{spoon\_powder} and \texttt{pour\_milk}.
In these cases, the model typically fails to predict the correct ordering.

\textbf{Actions per Video.}
\begin{table}
    \footnotesize
    \begin{tabularx}{0.48\textwidth}{Xrrr}
        \toprule
        cuts per video           & $ 4 $     & $ 2 $     & -         \\
        \midrule
        avg. \#actions per video & $ 12.5 $  & $ 25 $    & $ 50 $    \\
        midpoint hit             & $ 0.174 $ & $ 0.121 $ & $ 0.098 $ \\
        \bottomrule
    \end{tabularx}
    \caption{Different levels of video trimming for Cooking 2. More videos and less actions per video result in better performance.}
    \label{tab:cuts}
    \vspace{-0.5cm}
\end{table}
While our approach works well on Breakfast and Hollywood Extended, the results on Cooking 2
show its limitations. The dataset has many classes ($ 67 $) but only a small amount of
training videos ($ 220 $), which are very long and contain a huge amount of different actions.
These characteristics make it difficult for the multi-task learning to distinguish different classes, 
as many of them occur jointly in most training videos.
We show the importance of having enough videos by cutting each video of Cooking 2
into two/four parts (Table~\ref{tab:cuts}). This increases the number of videos and
reduces the number of actions per video. The more videos and the less actions
per video on average, the better are the results of our method.

\subsection{Inference given Action Sets}
\label{sec:givenActions}

\begin{table}[t]
    \centering
    \footnotesize
    \begin{tabularx}{0.48\textwidth}{Xrrr}
        \toprule
                                               & Breakfast           & Cooking 2             & Holl. Ext. \\
                                               & \textit{frame acc.} & \textit{midpoint hit} & \textit{jacc. idx} \\
        \midrule
        monte-carlo                            & $ 0.284 $  & $ 0.102 $             & $ 0.230 $            \\
        text-based                             & $ 0.280 $           & $ 0.106 $    & $ 0.242 $   \\
        \bottomrule
    \end{tabularx}
    \caption{Results of our method when the action sets are provided for inference.}
    \label{tab:givenActionSets}
    \vspace{-0.5cm}
\end{table}

So far, it has always been assumed that no weak supervision in form of action sets
is provided for inference. If the action sets for the videos are, for example, generated
using meta-tags of Youtube videos, however, they may as well be available during inference.
In this section, we evaluate our method under this assumption.

Let $ \mathcal{A} $ be the given action set for a video. During inference, only action sequences
that are consistent with $ \mathcal{A} $ need to be considered, \ie for a grammar
$ \mathcal{G} $, only sequences $ \mathbf{c}_1^N \in \mathcal{G} \cap \mathcal{A}^* $
are possible. If $ \mathcal{G} \cap \mathcal{A}^* $ is empty,
we consider all sequences $ \mathbf{c}_1^N \in \mathcal{A}^* $. 
The results are shown in Table~\ref{tab:givenActionSets}.
The above mentioned limitations on Cooking 2 again prevent our method from generating
a better segmentation. On Breakfast and Hollywood Extended, a clear improvement
of $ 5\% $ and $ 15\% $ compared to the inference without given action sets
(Table~\ref{tab:sota}) can be observed.


\section{Conclusion}

We have introduced a system for weakly supervised temporal action segmentation
given only unordered action sets. In contrast to ordered action sequences that have been
proposed as weak supervision by previous works, action sets are often publicly
available in form of meta-tags for videos and do not need to be annotated.  
Although action sets provide by far less supervision than ordered action sequences
and lead to a tremendously large search space, our method still achieves good
results.
Providing the possibility to incorporate data-driven grammars as well as
text-based information or human knowledge, our method can be adapted to
specific requirements in different video analysis tasks.

\textbf{Acknowledgements.}
The work has been financially supported by the DFG projects KU 3396/2-1 (Hierarchical Models for Action Recognition and Analysis in Video Data) and GA 1927/4-1 (DFG Research Unit FOR 2535 Anticipating Human Behavior) and the ERC Starting Grant ARCA (677650).
This work was supported by the AWS Cloud Credits for Research program.


{\small
\bibliographystyle{ieee}
\bibliography{references}
}

\end{document}